\documentclass[letterpaper]{article}

\usepackage{makeidx}         
\usepackage{graphicx}        
\usepackage{multicol}        
\usepackage[bottom]{footmisc}

\usepackage{times}
\usepackage{latexsym} 
\usepackage{theorem}
\usepackage{xspace}
\usepackage{amsmath}
\usepackage{mathtools}
\usepackage{amssymb}
\usepackage{booktabs}
\usepackage{paralist}
\usepackage{graphicx}
\usepackage{tikz}
\usepackage{pgfplots}
\usepackage{multirow}
\usepackage{wrapfig}
\usepackage{color,soul}
\newcommand*{\qed}{\hfill\ensuremath{\square}}
\pgfplotsset{compat=1.13}

\newtheorem{lemma}{Lemma}

\newtheorem{theorem}{Theorem}

\newtheorem{remark}{Remark}
\newtheorem{problem}{Problem}
\newtheorem{example}{Example}

\newcommand{\goal}{G}
\newcommand{\game}{P}
\newcommand{\states}{S}
\newcommand{\memstates}{M}
\newcommand{\act}{\mathcal{A}}
\newcommand{\trans}{\delta}
\newcommand{\obs}{\mathcal{Z}}
\newcommand{\obsadd}{\mathcal{Z}_A}
\newcommand{\obsmap}{\mathcal{O}}
\newcommand{\distr}{\mathcal{D}}
\newcommand{\initd}{I}
\newcommand{\supp}{\mathsf{Supp}}
\newcommand{\set}[1]{\{#1\}}

\newcommand{\Reach}{\mathsf{Reach}}
\newcommand{\uniform}{\mathsf{Uniform}}
\newcommand{\straa}{\sigma}
\newcommand{\prb}{\mathbb{P}}
\newcommand{\msize}{\mu}
\newcommand{\osize}{\nu}

\newcommand{\target}{T}
\newcommand{\obspart}{\obsmap_\bot}
\newcommand{\path}{\mathsf{Path}}
\newcommand{\nat}{\mathbb{N}}
\newcommand{\wt}{\widetilde}

\newcommand{\val}{\mathsf{Val}}
\usetikzlibrary{fit}
\usetikzlibrary{backgrounds}
\usetikzlibrary{shapes}
\usetikzlibrary{matrix}
\usetikzlibrary{patterns}
\usetikzlibrary{calc,arrows,automata}
\tikzstyle{State}=[circle, thick, minimum size=0.6cm, inner sep=0cm,draw=black]
\tikzstyle{BState}=[circle, thick, minimum size=0.8cm, inner sep=0cm,draw=black]
\tikzstyle{RState}=[circle, very thick, minimum size=0.8cm, inner sep=0cm,draw=red]

\begin{document}

\title{Sensor Synthesis for POMDPs with Reachability Objectives\thanks{The research was partly supported by AFRL FA8650-15-C-2546, ONR N00014-16-1-2051, DARPA W911NF-16-1-0001, Austrian Science Fund (FWF) Grant No P23499-N23, FWF NFN Grant No S11407-N23 (RiSE), ERC Start grant (279307: Graph Games), and Microsoft faculty fellows award.}}
\author{Krishnendu Chatterjee \\IST Austria \and Martin Chmel\'ik \\IST Austria \and Ufuk Topcu \\ University of Texas at Austin}
\date{}
%
%
\maketitle

\abstract{Partially observable Markov decision processes (POMDPs) are widely
used in probabilistic planning problems in which an agent interacts
with an environment using noisy and imprecise sensors. We study a
setting in which the sensors are only partially defined and the goal
is to synthesize ``weakest" additional sensors, such that in the
resulting POMDP, there is a small-memory policy for the agent that
almost-surely (with probability~1) satisfies a reachability objective.
We show that the problem is NP-complete, and present a symbolic
algorithm by encoding the problem into SAT instances. 
We illustrate trade-offs between the amount of memory of the policy 
and the number of additional sensors on a simple example. 
We have implemented our approach and consider three classical
POMDP examples from the literature, and show that in all the 
examples the number of sensors can be significantly decreased 
(as compared to the existing solutions in the literature)
without increasing the complexity of the policies.}

\section{Introduction}

In this work we study synthesis of sensor requirements for partially defined POMDPs, i.e., required precision of sensors, need for additional sensors, minimal set of necessary sensors, etc.  

\smallskip\noindent{\em POMDPs.}
\emph{Markov decision processes (MDPs)} are a standard model for 
systems that have both probabilistic and nondeterministic
behaviors~\cite{Howard}, and they provide a framework to
model and solve control and probabilistic planning problems~\cite{FV97,Puterman}.
The various choices of control actions for the controller (or planner) 
are modeled as {\em nondeterminism} while the stochastic response 
to the control actions is represented by the probabilistic behavior. 
In \emph{partially observable MDPs (POMDPs)} to resolve the nondeterministic 
choices in control actions the controller observes the state space 
according to observations, i.e., the controller can only view the observation of the 
current state, but not the precise 
state~\cite{PT87}.
POMDPs are a widely used model for several applications and research fields, such as in 
computational biology~\cite{Bio-Book}, speech processing~\cite{Mohri97}, 
image processing~\cite{IM-Book}, software verification~\cite{CCHRS11}, 
robot planning~\cite{kaelbling1998planning}, 
reinforcement learning~\cite{LearningSurvey}, to name a few. 

\smallskip\noindent{\em Reachability objectives.}
One of the most basic objectives is the \emph{reachability objective}, where given a set of target states, 
the objective requires that some state in the target set is visited at least once. 
The classical computational questions for POMDPs with reachability objectives 
are as follows: (a)~the \emph{quantitative} question asks for the existence 
of a policy (that resolves the choice of control actions) that ensures
the reachability objective with probability at least $0<\lambda\leq 1$; 
and (b)~the \emph{qualitative} question is the special case of the 
quantitative question with $\lambda=1$ (i.e., it asks that the objective
is satisfied almost-surely).

\smallskip\noindent{\em Previous results.} 
The quantitative question for POMDPs with reachability objectives 
is undecidable~\cite{PazBook} (and the undecidability result even holds for any 
approximation~\cite{MHC03}).
In contrast, the qualitative question is EXPTIME-complete~\cite{CDH10a,BGB12}. 
The main algorithmic idea to solve the qualitative question (that originates 
from~\cite{CDHR06}) is as follows: first construct the belief-support MDP 
explicitly (which is an exponential-size perfect-information MDP where every 
state is the support of a belief), and then solve the qualitative analysis 
on the perfect-information MDP (which is in polynomial time~\cite{CJH03,CH14,CH11}).
This gives an EXPTIME upper bound for the qualitative analysis of POMDPs, 
and a matching EXPTIME lower bound has been established in~\cite{CDH10a}.

\smallskip\noindent{\em Modeling and analysis.} 
In the design of systems there are two crucial phases, namely, the {\em modeling} phase,
where a formal model of the system is constructed, and the {\em analysis} phase,
where the model is analyzed for correctness. 
Currently POMDPs are typically used in the analysis phase, where in the modeling phase 
a {\em fully specified} POMDP for the system is constructed, which is
analyzed (in the model-checking terminology this is called {\em a posteriori} 
analysis or verification).
However, POMDPs are seldom used in the modeling phase, where the model 
is not yet fully specified.

\smallskip\noindent{\em Partially specified POMDPs.}
In this work we consider the problem in which a POMDP is {\em partially specified}
and can be used also in the modeling phase (i.e., {\em a priori} verification). 
To motivate our problem consider the standard applications in robotics or planning,
where the state space of the POMDP is obtained from valuations of the variables
of the system, and the sensors are designed to obtain the observations.
We consider a partially specified POMDP where the state space and the transitions 
are completely specified, but the observations are not. 
This corresponds to scenarios where (i)~the state space of the system is designed 
but the sensors have not yet been designed~\cite{C15,MDR2015} or (ii) the sensors are 
designed and there is a possibility to augment and annotate the state space, in order to
make the task for the agent simpler. 
In both scenarios the goal is to synthesize the observations (that is from the partially specified
POMDP obtain a fully specified POMDP) such that in the resulting POMDP there is a 
policy that satisfies the reachability objective almost-surely. 
Since additional sensors increase complexity, one goal is to obtain as few additional observations 
as possible; and since policies represent controllers another goal is to ensure
that the resulting policies are not too complex~\cite{ABZ10}. 
Concretely, we consider the following problem: given a partially specified
POMDP (where the observations are not completely specified), the problem 
asks to synthesize at most $\osize$ additional observations such that in the 
resulting POMDP there is a policy with memory size at most $\msize$ to ensure that 
the reachability objective is satisfied almost-surely.
Note that the problem we consider provides trade-offs between the 
additional observations (i.e., $\osize$) and the memory of the policy (i.e., $\msize$).

\smallskip\noindent{\em Significance of qualitative question.}
The qualitative question is of great importance as in several applications it 
is required that the correct behavior happens with probability~1.
For example, in the analysis of randomized embedded schedulers, the important 
question is whether every thread progresses with probability~1.
Moreover, though it might be sufficient that the correct behavior arises 
with probability at least $\lambda<1$, the correct choice of the threshold $\lambda$ 
is still challenging, due to simplifications and imprecisions introduced during 
modeling.  
Importantly it has been shown recently~\cite{CCGK16} that for the fundamental 
problem of minimizing the total expected cost to reach the target 
set~\cite{Bertsekas95,BG09,KMWG11,KMW12} under positive cost functions 
(or the stochastic shortest path problem), 
it suffices to first compute the almost-sure winning set, and 
then apply any finite-horizon algorithm for approximation.
Moreover, the qualitative analysis problem has also a close connection with 
planning: while the qualitative analysis problem is different as compared 
to strong or contingent planning~\cite{MBKS14,CPRT03,APG09}, 
it is equivalent to the strong cyclic planning problem~\cite{CPRT03,BCP06}.
Thus results for qualitative analysis of POMDPs carry over to strong cyclic planning.
Finally, besides the practical relevance, almost-sure convergence, 
like convergence in expectation, is a fundamental concept in probability 
theory, and provides the strongest probabilistic guarantee~\cite{Durrett}.

\smallskip\noindent{\em Our contributions.}
Our main contributions are as follows.
First, we show that when $\osize$ and $\msize$ are constants, 
then the problem we consider is NP-complete.
Note that the unrestricted problem (without restrictions on $\osize$ and 
$\msize$) is EXPTIME-complete, because we can use observations of at most the 
size of the state space, and the general qualitative analysis problem of fully 
specified POMDPs is EXPTIME-complete.
Second, we present an efficient reduction of our problem to 
SAT instances. This results in a practical, symbolic algorithm for the 
problem we consider and state-of-the-art SAT solvers, from artificial intelligence as well as many other 
fields~\cite{lingeling,rintanen11,BCCFZ99}, can be used for our problem.
Then, we illustrate the trade-offs between the amount of 
memory of the policy and the number of additional sensors on a simple example. 
Finally, we present experimental results.
We consider three classical POMDP examples from the literature, and show that
in these examples the number of observations (hence the number of sensors in practice) 
can be significantly decreased as compared to the existing models in the literature, 
without increasing the memory size of the policies.
We report scalability results on three examples showing that our 
implementation can handle POMDPs with ten thousand states. 

\section{Preliminaries}
A probability distribution $f$ on a finite set $X$ is a function $f:X \to [0,1]$ such 
that $\sum_{x\in X} f(x)=1$, we denote by  $\distr(X)$ the set of 
all probability distributions on $X$ and by $\uniform(X)$ the uniform distribution over a finite set $X$.
For a distribution $f \in \distr(X)$ we denote by $\supp(f)=\set{x\in X \mid f(x)>0}$
the support of~$f$.

\smallskip\noindent \textbf{POMDPs.}
A \emph{Partially Observable Markov Decision Process (POMDP)} is defined as a
tuple $\game=(\states,\act,\trans,\obs,\obsmap,\initd)$ where
\begin{compactitem}
\item (i)~$\states$ is a finite set of states;
\item (ii)~$\act$ is a finite alphabet of \emph{actions};
\item (iii)~$\trans:S\times\act \rightarrow \distr(S)$ is a 
 \emph{probabilistic transition function} that given a state $s$ and an
 action $a \in \act$ gives the probability distribution over the successor 
 states, i.e., $\trans(s,a)(s')$ denotes the transition probability from 
 $s$ to $s'$ given action~$a$; 
 \item (iv)~$\obs$ is a finite set of \emph{observations}; 
 \item (v)~$\initd \in \states$ is the unique initial state;
 \item (vi)~$\obsmap:\states \rightarrow \distr(\obs)$ is a probabilistic \emph{observation function} that 
  maps every state to a probability distribution over observations.
%
\end{compactitem}

\noindent \textbf{Plays.}
A \emph{play} (or a path) in a POMDP is an infinite sequence $(s_0,a_0,s_1,a_1,s_2,a_2,\ldots)$ 
of states and actions such that $s_0 = \initd$ and, for all $i \geq 0$, we have $\trans(s_i,a_i)(s_{i+1})>0$.
We write $\Omega$ for the set of all plays.

\smallskip\noindent \textbf{Policies.}
A \emph{policy (or a strategy)} is a recipe to extend prefixes of plays. That is, a policy  
is a function 
$\straa: (\obs \cdot \act)^* \cdot \obs \rightarrow \distr(\act)$
that, given a finite 
history of observations and actions, selects a probability distribution 
over the actions to be played next.
We present an alternative definition of policies with finite memory for POMDPs. 

\smallskip\noindent \emph{Policies with Memory.}
A policy with memory is a tuple $\straa = (\straa_u, \straa_n, \memstates, m_0)$ with the following elements: 
\begin{compactitem}
\item~$\memstates$ is a finite set of \emph{memory elements}. 
\item~The function $\straa_n : \memstates \rightarrow \distr(\act)$ is the \emph{action selection function} that maps the current memory element to a probability distribution over actions.
\item~The function $\straa_u : \memstates \times \obs \times \act \rightarrow \distr(\memstates)$ is the \emph{memory update function} that, given the current memory element, the current observation and action, updates the memory element probabilistically. 
\item~The element $m_0 \in \memstates$ is the \emph{initial memory element}.
\end{compactitem}
We will say a policy has memory size $n$ if the number of memory elements is $n$, i.e., $\vert \memstates \vert = n$. 

\smallskip\noindent \textbf{Probability Measure.}
Given a policy $\straa$ and a starting state~$\initd$, the unique probability measure obtained given 
$\straa$ is denoted as $\prb_\initd^{\straa}(\cdot)$~\cite{Billingsley,LittmanThesis}.

\smallskip\noindent{\bf Reachability Objectives.}
Given a set $\target \subseteq \states$ of \emph{target} states, a \emph{reachability objective} in a POMDP $\game$ is a measurable set $\varphi \subseteq \Omega$ 
of plays defined as follows: 
$\Reach(\target) = \{ (s_0, a_0, s_1, a_1, s_2 \ldots) \in \Omega \mid 
\exists i \geq 0:  s_i \in \target\}$, i.e., the set of plays, such that a state from the set of \emph{target} states $\target$ is visited
at least once.

In the remainder of the paper, we assume that the set of target states consists of a single \emph{goal} state, i.e., $T = \{ \goal\} \subseteq \states$. This assumption is w.l.o.g. because it is always possible to add a state $\goal$ with transitions from all target states in $T$.
Note, that there are no costs or rewards associated with transitions.

\smallskip\noindent{\bf Almost-Sure Winning.}
A policy $\straa$ is \emph{almost-sure winning} for a POMDP $\game$ with a reachability objective $\Reach(T)$ iff $\prb_\initd^\straa(\Reach(T)) = 1$.  
In the sequel, whenever we refer to a winning policy, we mean an almost-sure winning policy.

\section{Partially Defined Observation Functions}
\label{sec:part}

Traditionally, POMDPs are equipped with a fully defined observation function $\obsmap: \states \rightarrow \distr(\obs)$ that assigns to every state
of the POMDP a probability distribution over observations.
In order to model the \emph{partially defined} observation function, we assume the input POMDP~$\game$ is given with \emph{partially defined} observation function
$\obspart: \states \rightarrow \distr(\obs \cup \set{\bot})$. The probability distributions in the range of the function $\obspart$ contain an additional symbol $\bot$, and whenever for a state
$s \in \states$ we have $\bot \in \supp(\obspart(s))$, we will say that the state~$s$ has observations only partially defined.

\smallskip\noindent\emph{Observation function completions.}
We say a fully defined observation function $\obsmap$ is a \emph{completion} of a partially defined observation function $\obspart : \states \rightarrow \distr(\obs \cup \set{\bot})$ (and write $\obspart \prec \obsmap$) if all of the following conditions are met:
\begin{compactenum}
\item There exists a set $\obsadd$ of additional observations and the observation function $\obsmap: \states \rightarrow \distr(\obs \cup \obsadd)$ maps the states only to the set of old observations $\obs$ and the newly added observations $\obsadd$, i.e., the observations are defined for all states.
\item The function $\obsmap$ agrees on assigned observations with $\obspart$, i.e., for all states $s \in \states$ and observations $z \in \obs$, we have $\obspart(s)(z)= \obsmap(s)(z)$.
\end{compactenum}

Intuitively, given a POMDP with a reachability objective and a partially defined observation function $\obspart$, 
Problem~\ref{prob:1} asks, whether there exists a completion not using more than $\osize$ additional observations 
such that in the resulting POMDP there exists an almost-sure winning policy not using more than $\msize$ memory elements. 
More formally we study:



\begin{problem}
\label{prob:1}
Given a POMDP $\game = (\states,\act,\trans,\obs,\obspart,\initd)$ with a reachability objective $\Reach(\target)$, and two integer parameters $\msize>0$  and $\osize \geq 0$, 
decide whether there exists a completion $\obspart \prec \obsmap$ using additional observations $\obsadd$ and 
an almost-sure winning policy $\straa = (\straa_u, \straa_n, \memstates, m_0)$ for the $\Reach(\target)$ objective in the POMDP
$\game' = (\states,\act,\trans,\obs \cup \obsadd,\obsmap,\initd)$, with  $\vert \obsadd \vert \leq \osize$ and $\vert \memstates \vert \leq \msize$.
\end{problem}

\begin{example}
Consider a POMDP depicted in Figure~\ref{fig:simple}. There are three states corresponding to the position of the agent on the grid. The agent starts in the leftmost grid cell, and tries to move to the rightmost grid cell, where a treasure is hidden. There are three deterministic actions available to the agent: \emph{move-left}, \emph{move-right}, and \emph{grab-treasure}. When the action \emph{grab-treasure} is played in the rightmost cell, the agents wins, if it is played in any other cell the agent loses. The remaining two movement actions move the agent in the corresponding directions, if the wall is hit the agent loses.
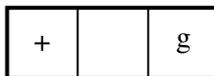
\begin{figure}[t]
\centering
\resizebox{0.25\linewidth}{!}{
\begin{tikzpicture}
  \tikzstyle{every node}=[font=\large]
  \draw [ultra thick] (0,0) rectangle (3,1);
  \draw[thick] (1,0)--(1,1);
  \draw[thick] (2,0)--(2,1);
  \node at (+0.5,+0.5) {+};
  \node at (+2.5,+0.5) {g};
\end{tikzpicture}
}
\caption{Grid POMDP}
\label{fig:simple}
\end{figure}

\begin{itemize}
\item In the setting where $\msize = 3$ and $\osize = 1$, the problem is satisfiable by a policy that plays actions in the following sequence \emph{move-right}, \emph{move-right}, and \emph{grab-treasure}.
\item In the setting where $\msize = 2$ and $\osize = 2$, the problem is satisfiable by an observation function that assigns the rightmost grid cell an observation different from the two remaining grid cells. The policy plays action \emph{move-right} in the first memory element until an observation corresponding to the rightmost cell is observed. After that it switches to the second memory element, where it plays action \emph{grab-treasure}.
\item In the setting where $\msize = 2$ and $\osize = 1$, the problem is not satisfiable, i.e., there is no two-memory almost-sure winning policy if all the states have the same observation.
\end{itemize}

\end{example}


\section{Complexity and SAT Encoding}
In this section we consider properties of almost-sure winning policies, the complexity of Problem~\ref{prob:1}, and 
its encoding to SAT instances.

\subsection{Complexity}

\begin{theorem}
\label{thm:np}
Deciding Problem~\ref{prob:1} given constant parameters~$\msize$ and~$\osize$ is NP-complete.
\end{theorem}

\noindent{\em Main ideas.} 
We remark that Theorem~\ref{thm:np} holds even if parameters $\msize$ and $\osize$ are polynomial in the size of the POMDP.
\begin{itemize}
\item {\em Inclusion in NP.} 
Note that for polynomial $\msize$ and $\osize$, a guess of the observation completion and 
the policy (if they exist) is polynomial. Thus we have polynomial-sized witnesses.
Given a policy and an observation function, we obtain a Markov chain where qualitative analysis is polynomial time
using standard discrete graph algorithms~\cite{CJH03,CH14,CH11}. 
Hence inclusion in NP follows.

\item {\em NP-hardness.} An NP-hardness result was established for a similar problem, namely, 
for no memory policies in fully specified two-player games with partial-observation, in~\cite[Lemma~1]{CKS13}.
The reduction constructed a game that is a DAG (directed acyclic graph), 
and replacing the adversarial player with a uniform distribution over choices shows 
that Problem~\ref{prob:1} is NP-hard even with $\msize=1$ (no memory policies) and $\osize=0$ 
(fully specified observation).
\end{itemize}


\subsection{SAT Encoding}
\label{subsec:sat}
In this section we present SAT encoding for Problem~\ref{prob:1},
which generalizes the special case of fully specified observation function 
studied in~\cite{CCD16}.

\smallskip\noindent{\em Standard Results.} We now present two basic lemmas.
The following lemma presents a standard result for qualitative analysis of POMDPs, and it basically follows from the fact that in a Markov chain 
for qualitative analysis, the exact probability distributions are not important, and the supports of the distributions  
completely characterize almost-sure winning.

\begin{lemma}
\label{lem:uniform}
Given an almost-sure winning policy $\straa = (\straa_u, \straa_n, \memstates, m_0)$ for a $\Reach(\target)$ objective, the policy $\straa'= (\straa'_u, \straa'_n, \memstates, m_0)$, where for $m \in \memstates$ the action selection function $\straa'_n$ is defined as $\straa'_n(m) = \uniform(\supp(\straa_n(m)))$, and for $m \in \memstates, a \in \act$, and $z \in \obs$ the memory update function $\straa'_u$ is defined as 
$$\straa'_u(m,z,a) = \uniform(\supp(\straa_u(m,z,a))),$$ 
and is also an almost-sure winning policy for $\Reach(\target)$.
\end{lemma}

Given a policy $\straa$ and a POMDP $\game=(\states,\act,\trans,\obs,\obsmap,\initd)$ and two state-memory pairs $(s,m), (s',m') \in \states \times \memstates$ we define 
a predicate $\path_{k,\straa,\game}((s,m),(s',m'))$  to be $\mathsf{True}$ iff there exists a sequence of state-memory pairs $((s_1,m_1),(s_2,m_2),\ldots,$ $(s_j,m_j))$ of length $j$ where 
$0 < j \leq k$, such that $s = s_1, m = m_1, s' = s_j$, and $m' = m_j$, and for all $1 \leq i < j$ there exists an action $a_i \in \act$, observation $z_i \in \obs$, such that $\straa_n(m_i)(a_i) > 0$, $\trans(s_i, a_i)(s_{i+1}) > 0$, $\obsmap(s_{i+1})(z_i) > 0$, and $\straa_u(m_i,z_i,a_i)(m_{i+1}) > 0$. 
Let $R_{\game,\straa}$ be the set of all pairs $(s, m) \in \states \times \memstates$ such that $\path_{k,\straa,\game}((I,m_0),(s,m))$ 
is $\mathsf{True}$ for some $k \in \nat$.
The following lemma states that almost-sure winning policies are characterized by paths of bounded length to the goal state.

\begin{lemma}
\label{lem:winning}
A policy $\straa$ is almost-sure winning in a POMDP~$\game$ iff for every state-memory  pair $(s,m) \in R_{\game,\straa}$ the predicate $\path_{k,\straa,\game}((s,m),(G,m'))$ for some $m' \in \memstates$ and $k = |\states| \cdot |\memstates|$ is $\mathsf{True}$.
\end{lemma}

\smallskip\noindent{\em Consequences for the SAT encoding.}
The consequences of the presented lemmas for the SAT encoding are as follows: Lemma~\ref{lem:uniform} allows to encode only the supports of the probability distributions of the policy $\straa$, i.e., a boolean property whether an action (resp. a memory element) is present in the support of the distribution $\straa_n$ (resp. $\straa_u)$. Lemma~\ref{lem:winning} allows to characterize state-memory pairs $(s,m)$ that are almost-sure winning by encoding the boolean predicate 
$\path_{k,\straa,\game}$ that represents existence of paths to the goal state.

\smallskip\noindent{\em Notations.}
Given a POMDP $\game$, reachability objective $\Reach(\target)$, a bound on the number of memory elements $\msize$, a bound on the number of additional observations $\osize$, and a path length $k \leq |\states|\cdot |\memstates|$ which is a parameter related to the length of paths in the POMDP, we will define a formula in conjunctive normal form (CNF) $\Phi_{k,\msize,\osize}$ that for a sufficiently large parameter $k$, e.g., $k = |\states| \cdot |\memstates|$, will be satisfiable if and only if there exists completion of the observation function using no more than $\osize$ additional observations and an almost-sure winning policy with no more than $\msize$ memory elements, i.e., the associated instance of Problem~\ref{prob:1} is true. 
We define the set $\obsadd = \set{z_1, z_2, \ldots, z_{\osize}}$ of additional observations, and denote by $\obs' = \obs \cup \obsadd$ the disjoint union of the old observations in $\obs$ and the newly added observations in $\obsadd$.
We describe the CNF formula $\Phi_{k,\msize,\osize}$ by defining all of its Boolean variables, followed by the clausal constraints over those variables.

\smallskip\noindent \textbf{Boolean Variables.} We first introduce the variables. 
\begin{itemize}
\item We begin by encoding the action selection function $\straa_n$ of the policy $\straa$. We introduce a Boolean variable $A_{m,a}$ for each memory-state $m \in \memstates$ and action $a\in \act$ to represent that action $a$ is played with positive probability in memory state $m$, i.e., that $\straa_n(m)(a) > 0$ (see Lemma~\ref{lem:uniform}).

\item Next, we encode the memory update function $\straa_u$. We introduce a Boolean variable $M_{m, z, a, m'}$ for each pair of memory-states $m,m' \in \memstates$, observation $z\in \obs'$ and action $a \in \act$. If such a variable is assigned to True, it indicates that, if the current memory-state is $m$, the current observation is $z$, and action $a$ is played, then it is possible that the new memory-state is $m'$, i.e., $\straa_u(m,z,a)(m') > 0$ (see Lemma~\ref{lem:uniform}).

\item We encode the completion $\obsmap$ of the partially defined observation function $\obspart$. We introduce a variable $O_{s,z}$ for every state $s \in \states$ and observation $z \in \obs'$. The intuitive meaning is that the observation function completion $\obsmap$ assigns to state $s$ observation $z$ with positive probability.

\item Boolean variables $C_{i, m}$ for each state $i \in \states$ and memory state $m \in \memstates$ indicate which (state, memory-state) pairs are reachable by the policy. 

\item The variables $P_{i,m,j}$ for all $i \in \states$, $m \in \memstates$, and $0 \leq j \leq k$, correspond to the proposition that there is a path of length at most $j$ from $(i,m)$ to the goal state, that is compatible with the policy.
\end{itemize}
 
\noindent \textbf{Logical Constraints.} We introduce the following clause for each $m \in \memstates$ to ensure that at least one action is chosen with positive probability for each memory state (see Lemma~\ref{lem:uniform}):
\begin{displaymath}
\bigvee_{j \in \act}A_{m,j}.
\end{displaymath}

To ensure that the memory update function is well-defined, we introduce the following clause for each $m \in \memstates$, $a \in \act$ and $z \in \obs'$ (see Lemma~\ref{lem:uniform}): 
\begin{displaymath}
\bigvee_{m' \in \memstates}M_{m, z, a, m'}.
\end{displaymath}
  
To ensure that every state $i$ has at least one observation $z$ in the support of the observation function, we introduce the following clause for every state $i \in \states$:
\begin{displaymath}
\bigvee_{z \in \obs'}O_{i, z}.
\end{displaymath}

For every state $i \in \states$ and every $z \in \supp(\obsmap(i))$, we enforce the consistency by adding the clause:
\begin{displaymath}
O_{i, z}.
\end{displaymath}

For every state $i \in \states$, with observations fully defined, i.e., $\bot \not \in \supp(\obsmap(i))$, for every additional observation $z \in \obsadd$ we add the following clause:
\begin{displaymath}
\neg O_{i, z}.
\end{displaymath}

The following clauses ensure that the variables $C_{i,m}$ will be assigned True for all pairs $(i,m)$ that are reachable using the policy:
\begin{displaymath}
(C_{i,m} \wedge A_{m,a} \wedge O_{j,z} \wedge M_{m, z,a, m'}) \Rightarrow C_{j,m'}.
\end{displaymath}
Such a clause is defined for each pair $m,m' \in \memstates$ of memory-states, each pair $i,j \in \states$ of states, each observation $z \in \obs$ and each action $a \in \act$ such that $\trans(i, a)(j) >0$.  

Therefore, the fact that the initial state and initial memory element is reachable is enforced by adding the single clause
\begin{displaymath}
C_{I,m_0}.
\end{displaymath} 

We introduce the following unit clause for each $m \in \memstates$ and  $0\leq j \leq k$, which says that the goal state with any memory element is reachable from the goal state and that memory element using a path of length at most $0$:
\begin{displaymath}
P_{G,m,0}.
\end{displaymath} 

Next, we define the following binary clause for each $i \in \states$ and $m \in \memstates$ so that, if the  pair $(i,m)$ of a state and a memory element is reachable, then the existence of a path from $(i,m)$ to the goal state is enforced (see Lemma~\ref{lem:winning}):  
\begin{displaymath}
C_{i,m} \Rightarrow P_{i,m,k}.
\end{displaymath}

Finally, we use the following constraints to define the value of variables $P_{i,m,j}$ for all $i \in \states$, $m \in \memstates$, and $0 \leq j \leq k$  in terms of the chosen policy (see Lemma~\ref{lem:uniform} and definition of predicate $\path$).

\begin{displaymath}
P_{i,m,j} \iff 
 \bigvee_{a \in \act} \!\!\left[A_{m,a} \!\wedge\! \left(\bigvee_{\substack{m' \in \memstates, z \in \obs,\\
i' \in \states : \trans(i,a)(i') > 0}}  \left[O_{i',z} \wedge M_{m,z,a,m'} \wedge P_{i',m',j-1} \right]\right)\right] .
\end{displaymath}

The conjunction of all clauses defined above forms the CNF formula $\Phi_{k,\msize,\osize}$.

\begin{theorem}
The formula $\Phi_{k,\msize,\osize}$ for $k \geq |\states| \cdot \msize$ is satisfiable, iff there exists a completion of the observation function using no more than $\osize$ additional observations and an almost-sure winning policy
using no more than $\msize$ memory elements.
\end{theorem}
\smallskip\noindent\textbf{Proof}
[Proof sketch.]
\smallskip\noindent\emph{Satisfiable formula $\Rightarrow$ completion and a policy:}
If the formula $\Phi_{k,\msize,\osize}$ is satisfiable, the SAT solver outputs a valuation $v$ of the variables. The boolean variables $O_{s,z}$ that are true according to $v$ encode the completion of the observation function, variables $A_{m,a}$ encode the action selection function $\straa_n$, and $M_{m,z,a,m'}$ encode the memory update function $\straa_u$. The fact that the encoded policy is almost-sure winning follows from the clauses and lemmas~1 and~2.


\smallskip\noindent\emph{Completion and a policy $\Rightarrow$ satisfiable formula:} Given a completion of the observation function and an almost-sure winning policy $\straa$ we show how to construct a satisfying valuation $v$ for the formula $\Phi_{k,\msize,\osize}$. The completion of the observation function $\obspart$ gives the valuation for the $O_{s,z}$ variables, the action selection function $\straa_n$ for the $A_{m,a}$ variables, and the memory update function $\straa_u$ for the $M_{m,z,a,m'}$ variables. The valuation for 
the $C_{i, m}$ and $P_{i,m,j}$ is obtained by constructing $R_{\game,\straa}$ and examining which state-memory pairs are reachable and the shortest path to a goal state. $\qed$




\subsection{Partial Specification with Constraints}
In the previous section we have presented a SAT encoding for POMDPs with partially specified observation functions. In this section we discuss additional constraints that might be desirable and our encoding can be easily extended to handle these constraints.

\smallskip\noindent\textbf{Non-distinguishable states.} In many scenarios it might be the case that there are states that cannot be distinguished by any available sensors, i.e., the observation assigned to these states must necessarily be the same. This can be enforced by adding the following clause for any pair $j,j' \in \states$ of non-distinguishable states and all the observations $z \in \obs'$.

\begin{displaymath}
O_{j,z} \Leftrightarrow O_{j',z}.
\end{displaymath}

\smallskip\noindent\textbf{Distinguishable states.} In some scenarios it might be the case that two states cannot have the same observation. This can be enforced by adding the following clause for any pair $j,j' \in \states$ of states and all the observations $z \in \obs'$.

\begin{displaymath}
(O_{j,z} \wedge \neg O_{j',z}) \vee (\neg O_{j,z} \wedge O_{j',z}).
\end{displaymath}

\smallskip\noindent\textbf{Dependencies among observations.} Various dependencies among observations can be expressed. For example in a state~$i$ whenever an observation $z$ is observed with positive probability also observation $z'$ is observed with positive probability can be expressed by the following clause:
\begin{displaymath}
O_{i,z} \Rightarrow O_{i,z'}.
\end{displaymath}


\smallskip\noindent\textbf{Adding sensor variables.} 
Let $P$ be a POMDP with the set of observations $\obs = \{z_1, z_2, \ldots, z_n\}$ and observation function $\obsmap$. By adding a new sensor $C$, that receives values $\val(C) = \{c_1,c_2, \ldots, c_l\}$, the new set of observations in the modified POMDP $\wt{P}$ is $\wt{\obs}  = \obs \times \val(C)$ with observation function $\wt{\obsmap}$. This corresponds to increasing the observation dimensionality, rather than increasing cardinality. 
Our approach allows to synthesise observations in POMDP $\wt{P}$ as follows: 
\begin{compactitem}
\item We set the observations of all states to be undefined. 
\item We add constrains to the resulting formula as follows:  In POMDP $\wt{P}$ an observation $(z,c) \in \wt{\obs}$ for some $c \in \val(C)$ is received with positive probability in state~$i$, i.e., $\wt{\obsmap}(i)((z,c)) >0$ if and only if the observation $z \in \obs$ is received in state $i$ in the original POMDP~$P$ with positive probability, i.e., $\obsmap(i)(z)>0$.

\medskip
\begin{compactenum}
\item For every state $i \in S$ and observation $z \in \supp(\obsmap(i))$ we add the following clause:

\begin{displaymath}
\bigvee_{c \in \val(C)} O_{i,(z,c)}
\end{displaymath}

\item For every state $i \in S$ and observation $z \not \in \supp(\obsmap(i))$ we add the following constraint:

\begin{displaymath}
\bigwedge_{c \in \val(C)} \neg O_{i,(z,c)}
\end{displaymath}



\end{compactenum}
\end{compactitem}

\smallskip\noindent\textbf{Deterministic observations function.} 
For every state $s \in \states$ we introduce the following clause:
\begin{displaymath}
\sum_{z \in \obs'}O_{i, z}  = 1 \;\;(\text{ exactly one of } O_{i,z} \text{ for } z \in \obs' \text{ is true}).
\end{displaymath}

\begin{remark}
\label{rem:det}
Deterministic observation functions are a special case of probabilistic observation functions. Therefore, the number of observations in the deterministic case is an upper bound for the probabilistic case. However, a probabilistic observation function might require less observations.
\end{remark}

\section{Experimental Results}
In this section we present experimental results and evaluate our approach on several POMDP examples
that were published in the literature. We have implemented the encoding presented in Section~\ref{subsec:sat} as
a program in Python and use the MiniSAT SAT solver~\cite{minisat} on an Intel(R) Xeon(R)@ 3.50GHz CPU.


\begin{remark}
In our experimental results we consider the synthesis of deterministic observation functions. As mentioned in Remark~\ref{rem:det}, deterministic observation functions provide upper bound for the number of observations required by 
probabilistic observation functions. Thus synthesizing deterministic observation functions with fewer observations is the more challenging problem, which we consider to illustrate the effectiveness of our approach.
\end{remark}

We present our results first on a small, simple example to illustrate how various selections of the memory
bounds $\msize$ and additional observation bounds $\osize$ affect the computed policies and discuss the possible
trade-offs between the memory vs. observation budgets in Problem~\ref{prob:1}.


  \begin{table}
  \centering

  \begin{tabular}[ht]{l|l|l|l|l|r|l}
  \toprule
  Name & Grid & \# States & $\msize$ & $\osize$ & Time (s) & SAT \\
  \midrule
  Escape2 & $2\times 2$ & 19 &  5 & 5 & 0.18 & $\surd$ \\
  Escape3 & $3\times 3$ & 84 &  5 & 5 & 1.22 & $\surd$ \\
  Escape4 & $4\times 4$ & 259 &  5 & 5 & 5.69 &  $\surd$\\
  Escape5 & $5 \times 5$ & 628 & 5 & 5 & 19.31 &  $\surd$\\
  Escape6 & $6\times 6$ & 1299 &  5 & 5 & 52.65 &  $\surd$\\
  Escape7 & $7\times 7$ & 2404 & 5 & 5 & 131.77 &  $\surd$ \\
  Escape8 & $8\times 8$ & 4099 &  5 & 5 & 280.19 &  $\surd$ \\
  Escape9 & $9 \times 9$ & 6564 &  5 & 5 & 674.42 &  $\surd$ \\
  Escape10 & $10\times 10$ & 10003 &  5 & 5 & 1519.48 & $\surd$\\
  \bottomrule
  \end{tabular}
  \caption{Escape instances.}
  \label{table:escape}
  \end{table}


\begin{table}[ht]
\centering
\begin{tabular}[ht]{l|l|l|l|l|r|l}
\toprule
Name & Grid & \# States & $\msize$ & $\osize$ & Time (s) & SAT \\
\midrule
Hallway1 & $7\times 5$ & 38 &  2 & 2 & 0.22 & $\times$ \\
Hallway1 & $7\times 5$ & 38 &  3 & 2 & 0.55 & $\surd$ \\
Hallway2 & $11\times 9$ & 190 &  3 & 2 & 5.95 &  $\times$\\
Hallway2 & $11\times 9$ & 190 & 3 & 3 & 5.28 &  $\times$\\
Hallway2 & $11\times 9$ & 190 &  4 & 2 & 20.82 &  $\surd$\\
Hallway3 & $11\times 10$ & 226 & 3& 2 & 6.53 &  $\times$ \\
Hallway3 & $11\times 10$ & 226 &  3 & 3 & 7.33 &  $\times$ \\
Hallway3 & $11\times 10$ & 226 &  4 & 2 & 28.98 &  $\surd$ \\
\bottomrule
\end{tabular}
\caption{Hallway instances.}
\label{table:hallway}
\end{table}

\begin{table}
\centering
\begin{tabular}[ht]{l|l|l|l|r|l}
\toprule
Name & \# States & $\msize$ & $\osize$ & Time (s) & SAT \\
\midrule
RockSample4 &  351 &  2 & 2 & 2.43 & $\surd$ \\
RockSample5 &  909 &  2 & 2 & 18.14 & $\surd$ \\
RockSample6 & 2187 &  2 & 2 & 95.28 &  $\times$\\
RockSample6 & 2187 & 2 & 3 & 165.87 &  $\times$\\
RockSample6 & 2187 &  3 & 2 & 519.21 &  $\surd$\\
RockSample7  & 5049 & 2 & 2 & 565.49 &  $\times$ \\
RockSample7  & 5049 &  3 & 2 & 565.43 &  $\times$ \\
RockSample7  & 5049 &  3 & 3 & 5196.40 &  $\surd$ \\
\bottomrule
\end{tabular}
\caption{RockSample instances.}
\label{table:rocksample}
\end{table}

\subsection{Deterministic Hallway}
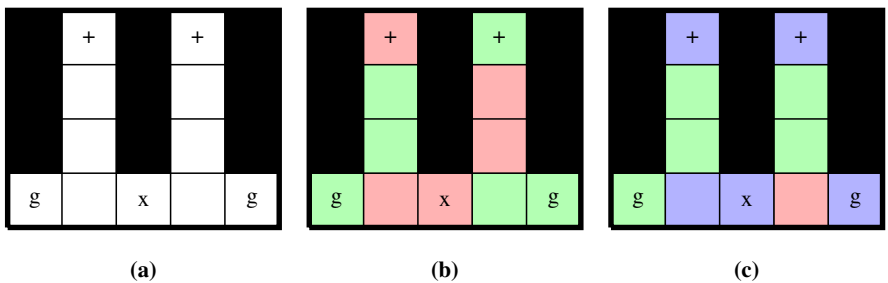
\begin{figure}[ht]
\resizebox{\linewidth}{!}{
\minipage{0.30\linewidth}
\resizebox{0.95\linewidth}{!}{
\begin{tikzpicture}
  \tikzstyle{every node}=[font=\large]
  \draw [fill=black] (0,1) rectangle (1,4);
  \draw [fill=black] (2,1) rectangle (3,4);
  \draw [fill=black] (4,1) rectangle (5,4);
  \draw[line width=0.1cm] (0,0)-- (5,0) -- (5,4) -- (0,4) -- (0,0);
  \draw[thick] (0,1)--(5,1);
  \draw[thick] (0,2)--(5,2);
  \draw[thick] (0,3)--(5,3);
  \draw[thick] (0,4)--(5,4);
  \draw[thick] (1,0)--(1,4);
  \draw[thick] (2,0)--(2,4);
  \draw[thick] (3,0)--(3,4);
  \draw[thick] (4,0)--(4,4);
  \node at (+1.5,+3.5) {+};
  \node at (+3.5,+3.5) {+};
  \node at (+0.5,+0.5) {g};
  \node at (+4.5,+0.5) {g};
  \node at (+2.5,+0.5) {x};
  \node at (+2.5,-0.8) {{\textbf{(a)}}};
\end{tikzpicture}
}
\endminipage
\hfill
\minipage{0.30\linewidth}
\resizebox{0.95\linewidth}{!}{
\begin{tikzpicture}
  \tikzstyle{every node}=[font=\large]
  \draw [fill=black] (0,1) rectangle (1,4);
  \draw [fill=black] (2,1) rectangle (3,4);
  \draw [fill=black] (4,1) rectangle (5,4);
  \draw [fill=green, opacity=0.3] (0,0) rectangle (1,1);
  \draw [fill=red, opacity=0.3] (1,0) rectangle (3,1);
  \draw [fill=green, opacity=0.3] (3,0) rectangle (5,1);
  \draw [fill=green, opacity=0.3] (1,1) rectangle (2,3);
  \draw [fill=red, opacity=0.3] (1,3) rectangle (2,4);
  \draw [fill=green, opacity=0.3] (3,3) rectangle (4,4);
  \draw [fill=red, opacity=0.3] (3,1) rectangle (4,3);
  \draw[line width=0.1cm] (0,0)-- (5,0) -- (5,4) -- (0,4) -- (0,0);
  \draw[thick] (0,1)--(5,1);
  \draw[thick] (0,2)--(5,2);
  \draw[thick] (0,3)--(5,3);
  \draw[thick] (0,4)--(5,4);
  \draw[thick] (1,0)--(1,4);
  \draw[thick] (2,0)--(2,4);
  \draw[thick] (3,0)--(3,4);
  \draw[thick] (4,0)--(4,4);
  \node at (+1.5,+3.5) {+};
  \node at (+3.5,+3.5) {+};
  \node at (+0.5,+0.5) {g};
  \node at (+4.5,+0.5) {g};
  \node at (+2.5,+0.5) {x};
  \node at (+2.5,-0.8) {{\textbf{(b)}}};

\end{tikzpicture}
}

\endminipage
\hfill
\minipage{0.30\linewidth}

\resizebox{0.95\linewidth}{!}{
\begin{tikzpicture}
  \tikzstyle{every node}=[font=\large]
  \draw [fill=black] (0,1) rectangle (1,4);
  \draw [fill=black] (2,1) rectangle (3,4);
  \draw [fill=black] (4,1) rectangle (5,4);
  \draw [fill=green, opacity=0.3] (0,0) rectangle (1,1);
  \draw [fill=blue, opacity=0.3] (1,0) rectangle (3,1);
  \draw [fill=red, opacity=0.3] (3,0) rectangle (4,1);
  \draw [fill=blue, opacity=0.3] (4,0) rectangle (5,1);
  \draw [fill=green, opacity=0.3] (1,1) rectangle (2,3);
  \draw [fill=blue, opacity=0.3] (1,3) rectangle (2,4);
  \draw [fill=green, opacity=0.3] (3,1) rectangle (4,3);
\draw [fill=blue, opacity=0.3] (3,3) rectangle (4,4);
  \draw[line width=0.1cm] (0,0)-- (5,0) -- (5,4) -- (0,4) -- (0,0);
  \draw[thick] (0,1)--(5,1);
  \draw[thick] (0,2)--(5,2);
  \draw[thick] (0,3)--(5,3);
  \draw[thick] (0,4)--(5,4);
  \draw[thick] (1,0)--(1,4);
  \draw[thick] (2,0)--(2,4);
  \draw[thick] (3,0)--(3,4);
  \draw[thick] (4,0)--(4,4);
  \node at (+1.5,+3.5) {+};
  \node at (+3.5,+3.5) {+};
  \node at (+0.5,+0.5) {g};
  \node at (+4.5,+0.5) {g};
  \node at (+2.5,+0.5) {x};
  \node at (+2.5,-0.8) {{\textbf{(c)}}};

\end{tikzpicture}
}
\endminipage
}
\caption{(a) Deterministic Hallway POMDP. (b) Synthesized~$\obsmap$ for Hallway $\msize=4$ and $\osize=2$. (c) Synthesized $\obsmap$ for Hallway $\msize=3$ and $\osize=3$. }

\label{fig:hallway1}

\end{figure}

We consider a simplification of the well-known Hallway problem~\cite{LCK95}, where an agent navigates itself
on a rectangular grid (see Figure~\ref{fig:hallway1}a).
There are four actions $N$, $E$, $S$, and $W$ available to the agent. For simplicity, 
all the movement on the grid is deterministic (probabilistic movement is considered in the Hallway problem later in 
the scalability evaluation). There are multiple initial states (depicted as $+$ in Figure~\ref{fig:hallway1}a) and the agent starts in any of them with uniform probability. The objective of the agent is to reach any of the goal states (depicted as $g$ in Figure~\ref{fig:hallway1}a). Whenever an agent
hits a wall or enters a trap state (depicted as $x$ in Figure~\ref{fig:hallway1}a) an absorbing state is reached, from which it is no 
longer possible to reach the desired goal states. We consider that there are no observations defined in the POMDP, i.e., for all states $s \in \states$ we
have $\obspart(s) = \bot$.

\smallskip\noindent\textbf{$4$ memory elements and $2$ observations.} In the setting, where $\msize=4$ and $\osize=2$ the SAT solver reports
that there exists a completion of $\obspart \prec \obsmap$  and an almost-sure winning policy $\straa$. We depict the synthesized observation
function $\obsmap$ in Figure~\ref{fig:hallway1}b, where the red color corresponds to the new synthesized observation $z_1$ and green color corresponds to
new synthesized observation $z_2$. The synthesized policy $\straa$ uses four memory elements $\memstates = \set{m_1,m_2,m_3,m_4}$. The synthesized action selection
function is defined as $\straa_n(m_1) = E, \straa_n(m_2) = W, \straa_n(m_3) = S, \straa_n(m_4) = S$. The computed policy initially updates its memory element to 
$m_3$ in case the first observation is $z_1$ (red) and to $m_4$ if the observation is $z_2$ (green), i.e., the information whether the agent starts in the 
left or right start state is stored in the memory element. Then action $S$ is played until the bottom row is reached (this is detected by changed observations,
from $z_2$ to $z_1$ in the left part, and from $z_1$ to $z_2$ in the right part). Finally, in case the agent is in the left part, memory element $m_3$ is updated
to $m_2$ and by action $W$ the goal state is reached. Similarly, in the right part, memory element $m_4$ is updated to $m_1$ and by action $E$ the goal state is reached.

\smallskip\noindent\textbf{$3$ memory elements and $3$ observations.} 
 In the setting, where $\msize=3$ and $\osize=3$ the SAT solver reports
there exists a completion of $\obspart \prec \obsmap$  and an almost-sure winning policy $\straa$. This allows to 
reduce the number of memory elements needed, provided we add one more observation to the POMDP.
We depict the synthesized observation
function $\obsmap$ in Figure~\ref{fig:hallway1}c, where red color corresponds to the new synthesized observation $z_1$, green color to
the new observation $z_2$, and blue color corresponds to the new observation $z_3$.
The synthesized policy $\straa$ uses three memory elements $\memstates = \set{m_1,m_2,m_3}$. The synthesized next-action selection
function is defined as $\straa_n(m_1) = W, \straa_n(m_2) = E, \straa_n(m_3) = S$. The computed policy starts with the initial memory
element $m_3$ and plays actions $S$ until either observation $z_1$ or $z_3$ is received. In case $z_1$ (red) is observed,
the policy is updated to memory element $m_2$ and reaches the goal state with action $E$. In case $z_3$ (blue) is observed,
the policy is updated to memory element $m_1$ and reaches the goal state with action~$W$.

\smallskip\noindent\textbf{$3$ memory elements and $2$ observations.}
In the setting, where $\msize=3$ and $\osize=2$ the SAT solver reports
there does not exist a completion of $\obspart \prec \obsmap$  that would allow for a two-memory almost-sure winning policy. This follows from the fact that at least three
memory elements are necessary for actions $S,E,W$, i.e., with the restriction $\msize=3$, there are no available memory elements to store additional information. As the agent needs to avoid hitting into walls, a randomized action selection function cannot be used.
It follows that there can be at most one memory element $m$ such that $\straa_n(m) = S$. It follows easily that, with only two observations and one memory element for action $S$, it is not possible to detect that the agent is already present in the bottom row.

\subsection{Scalability Evaluation}
In this part we demonstrate the scalability of our approach on three well-known POMDP examples of varying sizes.  Our results show that in all cases the observations
considered in these examples from the literature are unnecessarily refined and significantly less precise observations suffice even without making the policies more complicated.


\smallskip\noindent{\bf Escape POMDPs.}
The problem is originally based on a case study published in~\cite{SCL15}, where the goal is 
to compute a policy to control a robot in an uncertain environment.
A robot navigates on a square grid. There is an agent moving over the grid, 
and the robot must avoid being captured by the agent forever. 
The robot has four actions: move north, south, east, or west. 
These actions have deterministic effects, i.e., they always succeed. In the original POMDP instance, there are $179$ different observations.


\pgfplotsset{
every tick label/.append style={scale=0.8},
every axis/.append style={scale=0.5},
}

\begin{figure}[ht]

\centering

\begin{tikzpicture}
	\begin{semilogyaxis}[ylabel=$\#$ Observations ($\osize$), xlabel= $\#$ Memory ($\msize$)]
	\addplot[color=blue,mark=*] coordinates {
		(5,179)
		(5,5)
		(8,4)
		(12,3)
		(16,3)
	};
	\end{semilogyaxis}%
\end{tikzpicture}%
\caption{Memory vs. observation trade-off for Escape2}
\label{fig:escape-tradeoff}
\end{figure}
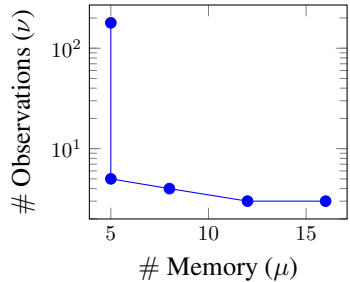

The memory and observation trade-offs for the smallest instance \emph{Escape2} are depicted on Figure~\ref{fig:escape-tradeoff}, which shows that
for $\msize=5$ and $\osize=179$ there exists an almost-sure policy. However, it is possible to significantly decrease the number of observations to 
$\osize=5$ and there is still an almost-sure winning policy with $\msize=5$. If $\msize$ is increased to~$8$, it is possible to decrease $\osize$ to $4$. If
the memory size $\msize$ is further increased to $12$, it is possible to reduce the number of observations $\osize$ to $3$. We illustrate the scalability results
in Table~\ref{table:escape}, where we report the number of states, the parameters $\msize,\osize$, the running time of the SAT solver, and whether the formula is satisfiable.
In all the cases with $\msize=5$ and $\osize=5$ there exists an almost-sure policy and the sizes of the instances go up to $10000$ states. There are approx. $1.6 \times 10^8$ clauses in the largest instance.

\smallskip\noindent{\bf Hallway POMDPs.}
Hallway POMDP instances are inspired by the Hallway problem introduced in~\cite{LCK95} and used later in~\cite{S04,SS04,BG09,CCGK14}. 
In the Hallway POMDPs, a robot navigates on a rectangular grid. The grid has barriers through which the robot cannot move, as well as trap locations that destroy the robot. The robot must reach a specified goal location. The robot has three actions: move forward, turn left, and turn right. 
The actions may all fail with positive probability, in which case the robot's state remains unchanged. The state is therefore comprised of the robot's location in the grid, and its orientation. Initially, the robot is randomly positioned among multiple start locations. Originally, the POMDP instances contain $16$ different observations. The results are reported in Table~\ref{table:hallway}, where we consider three different Hallway instances and vary the parameters $\msize$ for the number of memory elements and $\osize$ for the number of additional observations. For every entry we report the number of states, the parameters $\msize,\osize$, the running time of the SAT solver, and whether the formula is satisfiable. The results show that in all cases two observations are sufficient. In the smallest instance memory of size $3$ is sufficient. For larger instances, memory size needs to be increased to $4$. There are approx. $1.1 \times 10^7$ clauses in the largest instance.

\smallskip\noindent{\bf RockSample POMDPs.}
We consider a variant of the RockSample problem introduced in~\cite{SS04} and used later in~\cite{BG09,CCGK14}. 
The RockSample instances model rover science exploration.  
Only some of the rocks have a scientific value, and we will call these rocks ``good". 
Whenever a bad rock is sampled the rover is destroyed and a losing absorbing state is reached. 
If a rock is sampled for the second time, then with probability $0.5$ the action has no effect. 
With the remaining probability the sample is destroyed and the rock needs to be sampled one more time.
An instance of the RockSample problem is parametrized with a parameter $[n]$: $n$ is the number 
of rocks on a grid of size $3 \times 3$. The goal of the rover is to obtain two samples of good rocks. 
The results are presented in Table~\ref{table:rocksample}. Originally, the POMDP instances contain $15$ different observations.
 The results show that, with increasing sizes of the POMDP instances, either
increasing the memory size or increasing the number of additional observations is enough to obtain
an almost-sure winning policy. There are approx. $9.8 \times 10^7$ clauses in the largest instance.

\section{Conclusion}
In this work we consider POMDPs with partially specified 
observations, and the problem to synthesize additional 
observations along with small-memory almost-sure winning 
policies. 
Interesting directions of future work would be to consider
(a)~other aspects of partial specifications (such as transitions),
and (b)~other objectives, such as discounted-sum.

\bibliographystyle{plain}
\bibliography{pomdp}

\begin{thebibliography}{10}

\bibitem{APG09}
A.~Albore, H.~Palacios, and H.~Geffner.
\newblock A translation-based approach to contingent planning.
\newblock In {\em IJCAI}, pages 1623--1628, 2009.

\bibitem{ABZ10}
C.~Amato, D.S. Bernstein, and S.~Zilberstein.
\newblock {Optimizing fixed-size stochastic controllers for POMDPs and
  decentralized POMDPs}.
\newblock {\em AAMAS}, 21(3):293--320, 2010.

\bibitem{BGB12}
C.~Baier, M.~Gr{\"o}{\ss}er, and N.~Bertrand.
\newblock Probabilistic omega-automata.
\newblock {\em J. ACM}, 59(1), 2012.

\bibitem{BCP06}
P.~Bertoli, A.~Cimatti, and M.~Pistore.
\newblock Strong cyclic planning under partial observability.
\newblock {\em ICAPS}, 141:580, 2006.

\bibitem{Bertsekas95}
D.P. Bertsekas.
\newblock {\em Dynamic Programming and Optimal Control}.
\newblock Athena Scientific, 1995.
\newblock Volumes I and II.

\bibitem{lingeling}
A.~Biere.
\newblock Lingeling, plingeling and treengeling entering the {SAT} competition
  2013.
\newblock In {\em SAT Comp.}, 2013.

\bibitem{BCCFZ99}
A.~Biere, A.~Cimatti, E.M. Clarke, M.~Fujita, and Y.~Zhu.
\newblock Symbolic model-checking using {SAT} procedures instead of {BDD}s.
\newblock In {\em DAC}, pages 317--320, 1999.

\bibitem{Billingsley}
P.~Billingsley, editor.
\newblock {\em Probability and {M}easure}.
\newblock Wiley-Interscience, 1995.

\bibitem{BG09}
B.~Bonet and H.~Geffner.
\newblock Solving {POMDPs: RTDP-B}el vs. point-based algorithms.
\newblock In {\em IJCAI}, pages 1641--1646, 2009.

\bibitem{C15}
A.~Censi.
\newblock {A Mathematical Theory of Co-Design}.
\newblock {\em arXiv preprint arXiv:1512.08055}, 2015.

\bibitem{CCHRS11}
P.~Cern{\'y}, K.~Chatterjee, T.~A. Henzinger, A.~Radhakrishna, and R.~Singh.
\newblock Quantitative synthesis for concurrent programs.
\newblock In {\em Proc. of CAV}, LNCS 6806, pages 243--259. Springer, 2011.

\bibitem{CCGK14}
K.~Chatterjee, M.~Chmelik, R.~Gupta, and A.~Kanodia.
\newblock {Qualitative Analysis of POMDPs with Temporal Logic Specifications
  for Robotics Applications}.
\newblock {\em ICRA}, 2015.

\bibitem{CCGK16}
K.~Chatterjee, M.~Chmelik, R.~Gupta, and A.~Kanodia.
\newblock {Optimal Cost Almost-sure Reachability in POMDPs}.
\newblock In {\em AI}, 2016.

\bibitem{CCD16}
K.~Chatterjee, M.~Chmelik, and J.Davies.
\newblock {A Symbolic SAT-based Algorithm for Almost-sure Reachability with
  Small Strategies in POMDPs}.
\newblock {\em CoRR}, abs/1511.08456 (AAAI 2016), 2015.

\bibitem{CDH10a}
K.~Chatterjee, L.~Doyen, and T.~A. Henzinger.
\newblock Qualitative analysis of partially-observable {M}arkov decision
  processes.
\newblock In {\em MFCS}, pages 258--269, 2010.

\bibitem{CDHR06}
K.~Chatterjee, L.~Doyen, T.A. Henzinger, and J.F. Raskin.
\newblock Algorithms for omega-regular games with imperfect information.
\newblock In {\em CSL'06}, pages 287--302. LNCS 4207, Springer, 2006.

\bibitem{CH11}
K.~Chatterjee and M.~Henzinger.
\newblock Faster and dynamic algorithms for maximal end-component decomposition
  and related graph problems in probabilistic verification.
\newblock In {\em SODA}. ACM-SIAM, 2011.

\bibitem{CH14}
K.~Chatterjee and M.~Henzinger.
\newblock Efficient and dynamic algorithms for alternating {B{\"{u}}chi} games
  and maximal end-component decomposition.
\newblock {\em J. {ACM}}, 61(3):15, 2014.

\bibitem{CJH03}
K.~Chatterjee, M.~Jurdzi{\'n}ski, and T.A. Henzinger.
\newblock Simple stochastic parity games.
\newblock In {\em CSL'03}, LNCS 2803, pages 100--113. Springer, 2003.

\bibitem{CKS13}
K.~Chatterjee, A.~K{\"{o}}{\ss}ler, and U.~Schmid.
\newblock Automated analysis of real-time scheduling using graph games.
\newblock In {\em HSCC'13}, pages 163--172, 2013.

\bibitem{CPRT03}
A.~Cimatti, M.~Pistore, M.~Roveri, and P.~Traverso.
\newblock Weak, strong, and strong cyclic planning via symbolic model checking.
\newblock {\em Artificial Intelligence}, 147(1):35--84, 2003.

\bibitem{IM-Book}
K.~Culik and J.~Kari.
\newblock Digital images and formal languages.
\newblock {\em Handbook of formal languages}, pages 599--616, 1997.

\bibitem{Bio-Book}
R.~Durbin, S.~Eddy, A.~Krogh, and G.~Mitchison.
\newblock {\em Biological sequence analysis: probabilistic models of proteins
  and nucleic acids}.
\newblock Cambridge Univ. Press, 1998.

\bibitem{Durrett}
R.~Durrett.
\newblock {\em Probability: Theory and Examples (Second Edition)}.
\newblock Duxbury Press, 1996.

\bibitem{minisat}
N.~E\'{e}n and N.~S\"{o}rensson.
\newblock An extensible {SAT}-solver.
\newblock In {\em Theory and Applications of Satisfiability Testing}, pages
  502--518, 2003.

\bibitem{FV97}
J.~Filar and K.~Vrieze.
\newblock {\em Competitive {Markov} Decision Processes}.
\newblock Springer-Verlag, 1997.

\bibitem{Howard}
H.~Howard.
\newblock {\em Dynamic Programming and {Markov} Processes}.
\newblock MIT Press, 1960.

\bibitem{kaelbling1998planning}
L.~P. Kaelbling, M.~L. Littman, and A.~R. Cassandra.
\newblock Planning and acting in partially observable stochastic domains.
\newblock {\em Artif. Intell.}, 101(1):99--134, 1998.

\bibitem{LearningSurvey}
L.~P. Kaelbling, M.~L. Littman, and A.~W. Moore.
\newblock Reinforcement learning: A survey.
\newblock {\em JAIR}, 4:237--285, 1996.

\bibitem{KMW12}
A.~Kolobov, Mausam, and D.S. Weld.
\newblock A theory of goal-oriented {MDPs} with dead ends.
\newblock In {\em UAI}, pages 438--447, 2012.

\bibitem{KMWG11}
A.~Kolobov, Mausam, D.S. Weld, and H.~Geffner.
\newblock {Heuristic search for generalized stochastic shortest path MDPs}.
\newblock In {\em ICAPS}, 2011.

\bibitem{LittmanThesis}
M.~L. Littman.
\newblock {\em Algorithms for Sequential Decision Making}.
\newblock PhD thesis, Brown University, 1996.

\bibitem{LCK95}
M.~L. Littman, A.~R. Cassandra, and L.~P Kaelbling.
\newblock Learning policies for partially observable environments: Scaling up.
\newblock In {\em ICML}, pages 362--370, 1995.

\bibitem{MHC03}
O.~Madani, S.~Hanks, and A.~Condon.
\newblock On the undecidability of probabilistic planning and related
  stochastic optimization problems.
\newblock {\em Artif. Intell.}, 147(1-2):5--34, 2003.

\bibitem{MBKS14}
S.~Maliah, R.~Brafman, E.~Karpas, and G.~Shani.
\newblock Partially observable online contingent planning using landmark
  heuristics.
\newblock In {\em ICAPS}, 2014.

\bibitem{MDR2015}
A.~Mehta, J.~DelPreto, and D.~Rus.
\newblock Integrated codesign of printable robots.
\newblock {\em Journal of Mechanisms and Robotics}, 7(2):021015, 2015.

\bibitem{Mohri97}
M.~Mohri.
\newblock Finite-state transducers in language and speech processing.
\newblock {\em Comp. Linguistics}, 23(2):269--311, 1997.

\bibitem{PT87}
C.~H. Papadimitriou and J.~N. Tsitsiklis.
\newblock The complexity of {M}arkov decision processes.
\newblock {\em Mathematics of Operations Research}, 12:441--450, 1987.

\bibitem{PazBook}
A.~Paz.
\newblock {\em Introduction to probabilistic automata (Computer science and
  applied mathematics)}.
\newblock Academic Press, 1971.

\bibitem{Puterman}
M.~L. Puterman.
\newblock {\em {Markov} Decision Processes}.
\newblock John Wiley and Sons, 1994.

\bibitem{rintanen11}
J.~Rintanen.
\newblock Planning with {SAT}, admissible heuristics and {A*}.
\newblock In {\em IJCAI}, pages 2015--2020, 2011.

\bibitem{SS04}
T.~Smith and R.~Simmons.
\newblock Heuristic search value iteration for {POMDPs}.
\newblock In {\em UAI}, pages 520--527. AUAI Press, 2004.

\bibitem{S04}
M.T.J. Spaan.
\newblock A point-based {POMDP} algorithm for robot planning.
\newblock In {\em ICRA}, volume~3, pages 2399--2404. IEEE, 2004.

\bibitem{SCL15}
M.~Svorenova, M.~Chmelik, K.~Leahy, H.~F. Eniser, K.~Chatterjee, I.~Cerna, and
  C.~Belta.
\newblock {Temporal Logic Motion Planning using POMDPs with Parity Objectives}.
\newblock In {\em HSCC}, 2015.

\end{thebibliography}

\end{document}